\pdfoutput=1
\documentclass[11pt]{article}
\usepackage[preprint]{acl}

\usepackage{times}
\usepackage{tabularray}
\usepackage{latexsym}
\usepackage{graphicx}
\usepackage{booktabs}
\usepackage{makecell}
\usepackage{multirow}
\usepackage{tabularx}
\usepackage{longtable}
\usepackage{listings}
\usepackage{lipsum}
\usepackage[many]{tcolorbox}
\PassOptionsToPackage{table,dvipsnames}{xcolor}
\usepackage[table]{xcolor}
\usepackage{colortbl}

\usepackage[T1]{fontenc}
\usepackage[utf8]{inputenc}

\usepackage{microtype}

\usepackage{inconsolata}

\title{CAMPHOR: Collaborative Agents for Multi-input Planning and High-Order Reasoning On Device}

\author{Yicheng Fu \thanks{Work done as an intern at Apple.} \\
  Stanford University \\\And
  Raviteja Anantha \\
  Apple \\\And
  Jianpeng Cheng \\
  Apple \\}

\begin{document}
\maketitle
\begin{abstract}
% While server-side Large Language Models (LLMs) excel in tool usage and agentic reasoning, deploying Small Language Models (SLMs) on devices presents challenges in optimizing for not only accuracy, but also latency and user privacy. We propose CAMPHOR, an on-device SLM multi-agent framework capable of tracking multimodal user inputs and reasoning over personal information locally. CAMPHOR features a hierarchical architecture, with a high-order reasoning agent that decomposes tasks and orchestrates sub-agents such as personal context retrieval, tool calling, and plan generation. We optimize model size through parameter sharing across agents, and reduce latency and memory by prompt compression. We introduce the CAMPHOR benchmark, a dataset of multi-agent trajectories focused on mobile assistant use-cases. Our results show that the fine-tuned SLM agents can outperform closed-source LLMs in task completion, while eliminating server-device communication costs and safeguarding user privacy.
While server-side Large Language Models (LLMs) demonstrate proficiency in function calling and complex reasoning, deploying Small Language Models (SLMs) directly on devices brings opportunities to improve latency and privacy but also introduces unique challenges for accuracy and memory. We introduce CAMPHOR, an innovative on-device SLM multi-agent framework designed to handle multiple user inputs and reason over personal context locally, ensuring privacy is maintained. CAMPHOR employs a hierarchical architecture where a high-order reasoning agent decomposes complex tasks and coordinates expert agents responsible for personal context retrieval, tool interaction, and dynamic plan generation. By implementing parameter sharing across agents and leveraging prompt compression, we significantly reduce model size, latency, and memory usage. To validate our approach, we present a novel dataset capturing multi-agent task trajectories centered on personalized mobile assistant use-cases. Our experiments reveal that fine-tuned SLM agents not only surpass closed-source LLMs in task completion F1 by~35\% but also eliminate the need for server-device communication, all while enhancing privacy.
\end{abstract}

\section{Introduction} % (Jianpeng)

Server-side Large Language Models (LLMs) are powerful semantic parsers that interpret user intent and map queries to executable function calls. To ground a query within a personal environment such as an open toolbox, retrieval-augmented generation (RAG)~\cite{borgeaud2022improving} can be adopted to pre-fill the LLM prompt with external knowledge relevant to the user query, such as top-K most relevant tools. An orthogonal strategy is long-context language modeling~\cite{beltagy2020longformer,zaheer2020big}, which pre-loads the prompt with all available external knowledge, taking advantage of a larger context window up to 128K tokens~\cite{dubey2024llama}.
However, a server-side LLM is not optimal for a mobile assistant due to privacy and latency concerns.

\textbf{Privacy}. 
User queries to mobile assistants are often ambiguous, making it crucial to ground them in personal information, such as contacts, installed tools, and past activities. While the assistant needs access to personal data to improve understanding, it must also prioritize user privacy by keeping sensitive information on the device. Even private cloud solutions are not ideal for this, as they commonly avoid storing user-specific data, which prevents KV caching in multi-turn dialogues~\cite{li2024personal}.

\textbf{Latency}. 
In addition to the limited flexibility of KV caches due to privacy concerns, server-side LLMs introduce extra latency between understanding (which occurs on the server) and execution (which happens on the user’s device). This latency can degrade the user experience, particularly for solutions requiring multiple server-device round trips. For instance, the ReAct framework~\cite{yao2023react}, which breaks down the understanding task into multiple steps and reasons over intermediate execution results, is hindered by this added latency.

A more effective solution to address latency and privacy concerns is to deploy a small language model (SLM) on-device, allowing it to process personal data and interpret user queries locally. However, SLMs introduce new challenges related to accuracy due to tighter budgets on prompts and KV caches. 
While a RAG approach is scalable across personal databases, the retriever is a separate model which gates the performance of the language model since it is impossible to achieve a perfect recall given the fixed prompt budget~\cite{fan2024survey}. The problem becomes worse for compositional queries which are naturally harder for retrieval. 
Consider the following example:
\begin{verbatim}
Can you show me the cheapest flight options
to Barcelona next month and add it to my ca-
lendar? Also, let my travel buddy know about
our trip plan.
\end{verbatim}
The query is both compositional and includes a personal entity (my travel buddy). Successfully retrieving the appropriate tool requires task decomposition, which depends on the granularity of the available toolbox (i.e., the tools present on the user’s device). While long-context modeling allows the language model to consider all relevant tools and personal entities during query parsing, it also rapidly exhausts the prompt budget, increases KV cache overhead, and introduces a "needle-in-a-haystack" challenge~\cite{liu2024lost} for the SLM.

We argue that a key to successful on-device understanding is an agentic approach that decomposes the understanding task in functional space, proactively seeks personal information and runtime feedback before generating the final function call.
Specifically, we propose CAMPHOR, an SLM-based collaborative agent understanding system for grounded query parsing. CAMPHOR features a hierarchical architecture, with a high-order reasoning agent that incrementally solves a query by orchestrating sub-agents. Each sub-agent is responsible for a unique functionality, such as retrieving personal entities, fetching latest user activities, and calling task-completion APIs. We apply tool compression techniques that enable the SLM to reason over device-specific toolboxes while satisfying prompt and memory budgets. Overall, CAMPHOR is designed to offer optimal accuracy and speed, while protecting user data on device.

\begin{figure}[ht]
    \centering
    \includegraphics[width=\columnwidth]{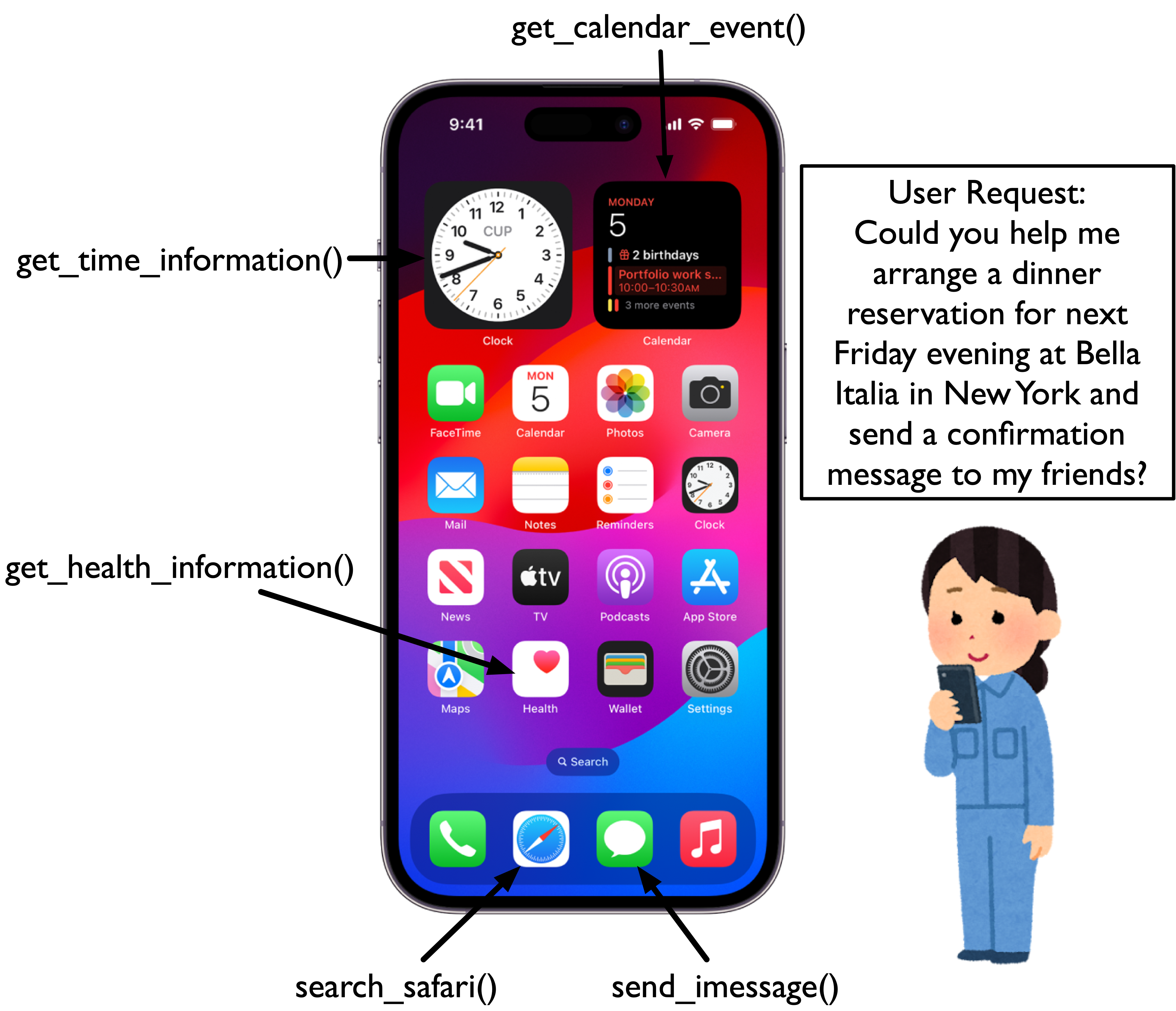} % Use \textwidth to scale across both columns
    \caption{CAMPHOR dataset simulates a user's smartphone environment, encompassing diverse personal information stored across multiple apps on the device.}
    \label{fig:personal}
\end{figure}

We release the CAMPHOR dataset\footnote{The dataset will be released publicly following the publication of this work.}, which consists of multi-agent execution trajectories focused on mobile assistant use cases (as shown in Figure~\ref{fig:personal}). The dataset is unique in how it simulates user device states and ground query understanding to personal information.
We fine-tune the SLM-based CAMPHOR agent on this dataset, and prove that fine-tuned SLMs can outperform closed-source LLMs in terms of task completion metrics, while eliminating server-device communication costs and safeguarding user privacy.

\section{Related Work}

\textbf{Small Language Models and On-device Agents} demonstrate benefits of faster inference, lower latency, and enhanced privacy protection. Studies show that SLMs like Mistral~\cite{jiang2023mistral}, Phi~\cite{abdin2024phi}, TinyLlama~\cite{zhang2024tinyllama}, MobileLLM~\cite{liumobilellm}, MiniCPM~\cite{hu2024minicpm}, and Gemma~\cite{team2024gemma}, when fine-tuned for specific tasks, can outperform prompting Large Language Models (LLMs). In particular for the function calling task, the Octopus series~\cite{chen2024octopus} has achieved remarkable accuracy, exceeding 97\% for function calling on device. Most related to our work is the on-device Octo-planner~\cite{chen2024octo}, which breaks down a query into multiple subqueries for function call generation. However, we argue that query decomposition in natural language space is an unconstrained optimization problem, as the granularity of decomposition depends on the available toolset.

\textbf{Multi-agent Planning} is the process where multiple agents, each with unique capabilities, knowledge, and objectives, work together towards shared or interrelated goals. The rise of large language models (LLMs) has significantly advanced the development of multi-agent planning, as tasks for each agent can potentially be solved through prompts. Agentic frameworks like ReAct~\cite{yao2023react}, Reflexion~\cite{shinn2024reflexion}, LATS~\cite{zhoulanguage}, SwiftSage ~\cite{lin2024swiftsage}, and AUTOACT~\cite{qiao2024autoact} continuously prompt LLMs to reflect on and critique their past actions, sometimes incorporating additional external information, such as environmental observations and feedback. In this work we focus on multi-agent that solves user queries while understanding user data on a device. Examples of such data includes past user actions, personal entities and installed toolsets~\cite{wu2024avataroptimizingllmagents}. 

\textbf{Retrieval Augmented Generation and Long-context Language Models} are two orthogonal approaches to ground a fine-tuned LM with external data sources, which in this work include the dynamic set of personal entities and tools~\cite{borgeaud2022improving}. A standard workflow of RAG includes possibly a query generation step (or a query decomposition step for compositional utterances)~\cite{ma2023query,rackauckas2024rag} followed by sparse and/or dense retrieval. The retrieval model is commonly a separate set of parameters which can be trained either separately or jointly with the LLM. A major limitation of RAG is that a sub-optimal retrieval model will gate the performance of the LLM which has access to more contextual information. In contrast, long-context LLMs allow for the direct incorporation of more external data into the prompt~\cite{beltagy2020longformer,zaheer2020big,kitaev2019reformer,ding2023longnet}. However, this comes at the cost of increased size of prompt and KV caches, making it impractical for small language models (SLMs) and on-device deployment.

\textbf{Prompt Compression} is an optimization to reduce the number of prompt tokens at least at the inference time.  We adopt the technique to enable SLMs to retrieve directly from a dynamic toolbox. Related to this work are the work of Gist tokens~\cite{mu2024learning}, Parallel Context Encoder~\cite{yen2024long}, and Squid which compresses a piece of long text into a single embedding~\cite{chen2024dolphin}. These approaches differ in terms of how the compressed embedding is learned and incorporated with the base LLM, as either prompt tokens or late fusions in the attention layer. 

\section{Methodology}
\subsection{Agents Overview}

%Let us use the following example \textit{delete camphor from my %LLM discussion and attach it to my team meeting} to motivate %CAMPHOR. The following describes how a RAG-based planner will %work:
%\begin{enumerate}
%    \item Generate the first sub-query \textit{delete camphor %from my LLM discussion}.
%    \item Retrieve K tools using the first sub-query.
%    \item Generate the first program for the first sub-query with the K tools in the prompt.
%    \item Generate the second sub-query \textit{add camphor to my team meeting}.
%    \item Retrieve K tools using the second sub-query.
%    \item Generate the second program for the second sub-query with the K tools in the prompt.
%\end{enumerate}

CAMPHOR is a collaborative agent framework that performs grounded query parsing on a user device. It consists of the following agents, including an orchestrator:
\begin{itemize}
  \item \textit{High-order reasoning agent} plans the process of understanding a user query, solving the query by determining the order in which other expert agents are invoked, effectively using the expert skills to complete sub-tasks.
\end{itemize}
and various experts:
\begin{itemize}
  \item \textit{Personal context agent} generates function calls to search relevant personal context that would be helpful in resolving entity ambiguities and under-specified queries. \textbf{The set of function calls that can be invoked by the agent is unique for each user device}, as the databases of personal entities are linked to the apps installed on a user device.
  \item \textit{Device information agent} generates generic function calls to retrieve device information including current location, time and screen entities. 
  \item \textit{User perception agent} represents a single function call to fetch the recent user activities on device. 
  \item \textit{External knowledge agent} generates generic function calls to seek information from external sources including web search, Wikipedia and calculator. 
  \item \textit{Task completion agent} generates function calls to represent the user intent towards task completion. \textbf{The set of function calls that can be invoked by the agent is unique for each user device}, as the capabilities are determined by the apps installed on a user device.
\end{itemize}

\begin{figure}[ht]
    \centering
    \includegraphics[width=\columnwidth]{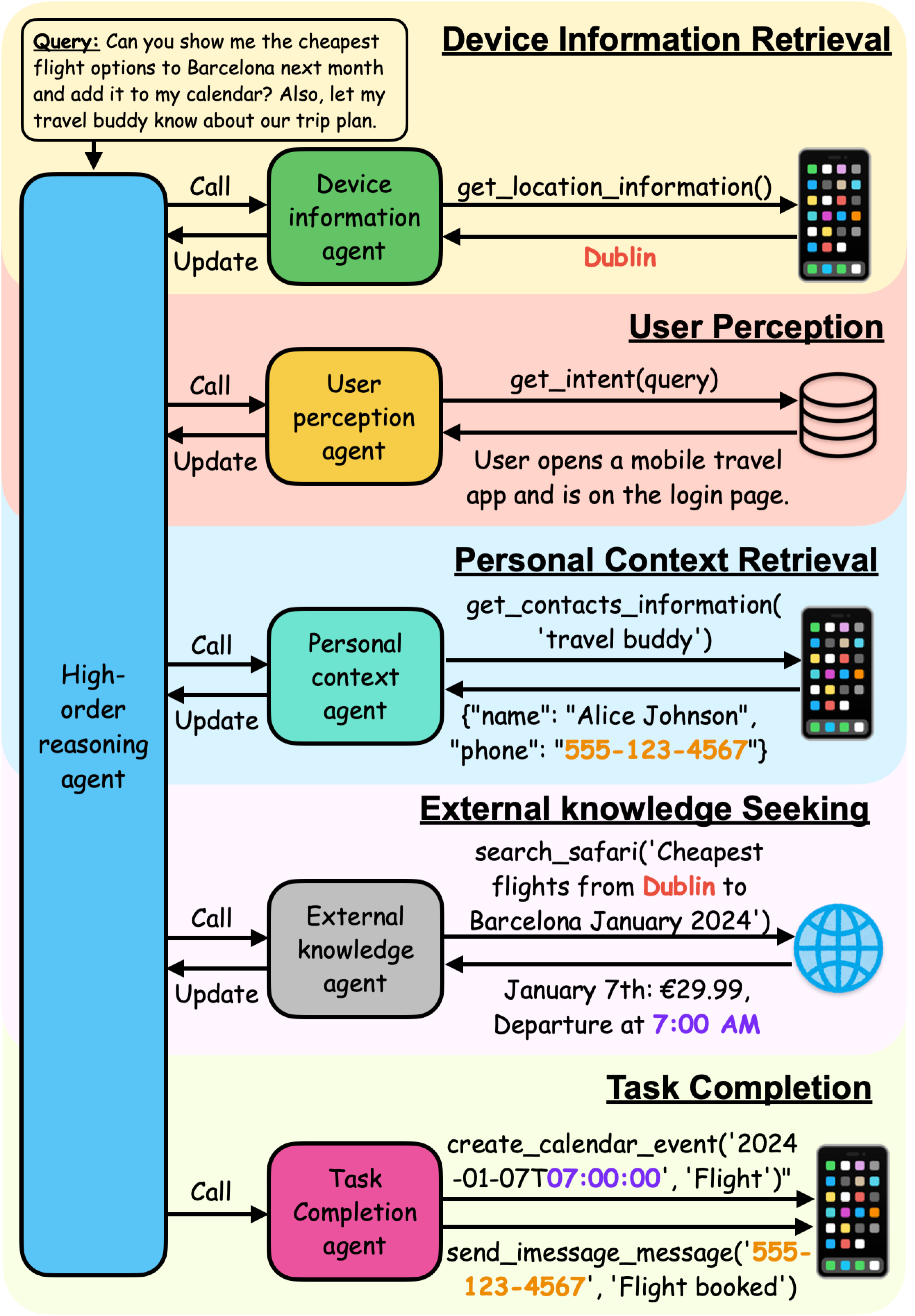} % Use \columnwidth to scale to 1 column
    \caption{An overview of multiple agents in CAMPHOR. The figure includes all agents for completeness. In practice, a subset of the agents can be invoked in arbitrary order until task completion.}
    \label{fig:framework}
\end{figure}

Take the following query as an example
\begin{verbatim}
Can you show me the cheapest flight options
to Barcelona next month and add it to my ca-
lendar? Also, let my travel buddy know about
our trip plan.
\end{verbatim}
The high-order agent proactively gathers personal information to understand the user intent. This includes \textit{Device information agent} to obtain the current location and \textit{Personal context agent} to look up the entity \textit{travel buddy}. The \textit{Task completion agent} is finally invoked to generate the task completion function calls. 
Figure~\ref{fig:framework} shows a high-level overview of all CAMPHOR agents.

%In addition, CAMPHOR optionally includes a reflection-and-correction loop:
%\begin{itemize}
%  \item \textit{Reflection agent} reviews the function calling results for the task completion agent. If the execution result is unsatisfying, the reflection agent will provide thoughts for re-planning, unless the execution result has to be feedback to prompt the user. 
%\end{itemize}

%JC to add example

\subsection{Dynamic Prompt Construction \label{dp}} % (Jianpeng, Yicheng)
We model all agents in CAMPHOR with the same underlying SLM.
A general formulation of all agents is that they take as input an agent-specific prompt and produces a function call which can be executed. The execution result of each expert agent is sent back to instruct the higher-order agent and the next expert agent. 
An agent prompt $p_a$ is generated by a template formatting function $f(i_a, h_a, t_a)$, where $i_a$ is an agent-specific task instruction and $h_a$ stands for the message history (i.e., the past agent actions and observations) that the agent has access to.

$t_a$ is an optional parameter representing function definitions that go into the prompt. Note that most agents actually use a static set of functions that are shared across  user devices. The static set of function calls and parameters can be directly memorized by the model without definitions revealed in the prompt.  However, two agents—the personal context agent and the task completion agent—interact with a dynamic set of device-specific functions. This is because the entity databases and capable tools are dependent on the apps installed on each user's device. For these two agents, we need to present a dynamic set of function definitions in the prompt.

\subsection{Prompt Compression \label{pc}} % (Jianpeng, Yicheng)
As discussed earlier and demonstrated through experiments in Section~\ref{rag}, a RAG-based approach is sub-optimal because the retriever gates the performance of an SLM. Additionally, it is impractical to include the entire set of function definitions in the prompt, as this would quickly exceed the prompt token limit. To address this, we compress each function definition into a single token, which is then appended to the beginning of the prompt. This prompt compression approach is reminiscent of the cross-modality token used in multi-modal language modeling. By doing so, the agent can still access and reason over the full set of function definitions, while significantly reducing the number of input tokens---by a factor corresponding to the average length of function definitions in the prompt.

\begin{figure*}
    \centering
    \includegraphics[width=\textwidth]{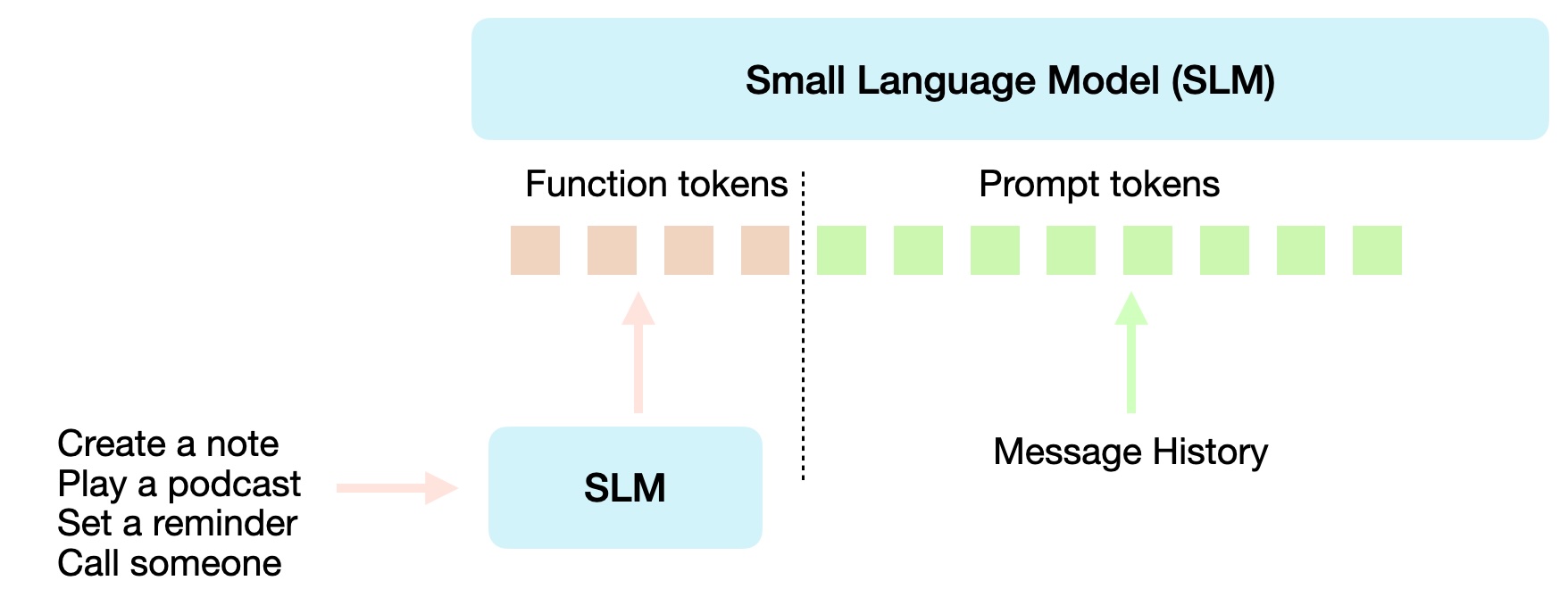} % Use \columnwidth to scale to 1 column
    \caption{Prompt compression technique. We use the pre-trained SLM itself as a text encoder to generate a single-token embedding for each function description, by taking the output embedding of the final token therein. The compressed function tokens are appended to the beginning of the prompt.}
    \label{fig:compress}
\end{figure*}

We opt for the SLM itself as a text encoder to obtain the single-token embedding for each function definition, by taking the output embedding of the last token therein, as illustrated in Figure~\ref{fig:compress}. The choice is motivated by the fact that the language model is already pre-trained to encode text, offering meta-learning generalization. During fine-tuning, gradients will not be back-propagated through the function tokens. Comparing to gist tokens~\cite{mu2024learning} which also leverage a pre-trained language model to encode texts as KV caches, our approach significantly reduces the cache size since only a single embedding is needed for each function definition, whose KV caches are computed on the fly of language model inference. 

\paragraph{Positional Embeddings.} We set custom position indices for the computation of the Rotary Positional Embeddings~\cite{su2024roformer}. Every function token in the prompt shares the same position index 0 while the first token in the formal prompt starts with with position index 1. Function tokens are restricted from attending to each other, but each prompt token can attend to all function tokens, reasoning over the toolbox jointly.

\section{CAMPHOR Dataset} % (Yicheng)
\label{sec:benchmark}
A central focus of CAMPHOR is personalized planning and query understanding on device. However, existing function calling datasets~\cite{patil2023gorilla,qintoolllm} only provide task completion annotations for user queries but not incorporating personal knowledge for understanding. On the other hand, there exist a few datasets on agent planning but they largely focus on mathematics~\cite{cobbe2021trainingverifierssolvemath, mishra2022lila, lumathvista} and common sense reasoning~\cite{talmor2019commonsenseqa, geva2021did} instead of query parsing.

To this end, we created the CAMPHOR dataset by annotating each query with a trajectory of function calls that demonstrate how a multi-agent system proactively fetches personal information to solve a user query by breaking down the understanding task into smaller actions. The dataset is developed by assigning a personal device state to each query, which includes a randomly sampled history of user activities, as well as the personal entities and tools available on the device. Each query in the dataset is generated by GPT-4o based on a device state and a set of global function definitions. The GPT-4o is also instructed to annotate the query in a multi-step fashion. The execution results are fetched from the device state for each function calling, which are then used to guide the next step of annotation. %This process is illustrated in Figure ~\ref{fig:data}. 
The final solution path is reviewed and verified with human oversight. Overall, the CAMPHOR dataset contains 3,410 queries, which are split into 2,728 for training and 682 for test. The dataset is flattened, resulting in 35,444 prompt-completion pairs for SLM fine-tuning, with an average of 10.39 pairs per query.

\section{Experiments} % (Yicheng)
\label{sec:experiments}

We consider two SLM candidates for fine-tuning the CAMPHOR agents: Phi-3.5 and Gemma-2. The sequence of prompt and completion pairs associated with each query is obtained by unrolling the ground truth trajectories in the dataset. 
%Table x shows an example of the unrolled prompts and completions. 

% \begin{table*}[ht]
% \centering
% \begin{tabular}{p{0.45\textwidth} p{0.45\textwidth}}{ll}
% \toprule
% Prompt & Completion \\
% \midrule
% [User] Call Jessica now. Also, find a podcast about improving communication skills and play it afterwards. \\
% [High Order Reasoning Agent]: & [Personal Context Retrieval Agent] \\
% \bottomrule
% \end{tabular}
% \caption{Prompt-completion pair example)}
% \end{table*}

\subsection{Evaluation Metrics}
%We propose five metrics to assess the performance of large language models (LLMs). The first two metrics evaluate the execution process, while the final three focus on the quality of the model’s final response.

We consider three end-to-end evaluation metrics on task completion:
\begin{itemize}
    %\item \textit{Personal Context Retrieval Recall} measures the proportion of functions that are successfully retrieve relevant personal contexts.
    %\item \textit{Fact Retrieval Recall} measures the proportion of FR API calls that successfully retrieve relevant facts.
    %\item \textit{Textual Response Quality} measures the degree of similarity between the system's textual response and the ground truth reference response.
    \item \textit{Tool F1} measures the accuracy of the function names used in task completion function calls. F1 is selected as the metric because it  not only accounts for true-positive predictions within the ground truth set, but also penalizes false-positive predictions outside of it.
    \item \textit{Delexicalized Plan F1} measures the accuracy of both function names and parameters in task completion function calls. A true-positive prediction must not include any parameter hallucinations. The prediction is measured at the abstract syntax tree level, disregarding the order of parameters.
    \item \textit{Plan F1} measures the accuracy of both function names, parameters and their values in task completion function calls. A subset of the parameters does support a open set of values, instead of closed-set enums.  We adopt a lenient match rule to evaluate open-ended values: a match is incurred if the Sentence-BERT~\cite{reimers2019sentence} embedding similarity between target and predicted values is higher than threshold 0.7.
\end{itemize}

\begin{table*}[ht]
\centering
\caption{Claude 3.5 Sonnet performance with different prompting strategies.}
\label{tab:claude}
\begin{tabular}{cccccc}
\toprule
\multirow{2}{*}{\textbf{Metric}} & \multicolumn{5}{c}{\textbf{Prompting Strategy}}  \\
\cmidrule(l){2-6}
& \textbf{CAMPHOR} & \textbf{AUTOACT} & \textbf{Reflexion} & \textbf{ReAct} & \textbf{Static} \\
\midrule
\textbf{Tool F1} / \% & \cellcolor{gray!20}\textbf{51.31} & 48.59 & 48.94 & 36.72 & 35.7 \\
\textbf{Delexicalized Plan F1} / \% & \cellcolor{gray!20}\textbf{30.07} & 25.77 & 24.49 & 19.16 & 18.78 \\
\textbf{Plan F1} / \% & \cellcolor{gray!20}\textbf{27.96} & 24.77 & 22.53 & 18.36 & 17.18 \\
\bottomrule
\end{tabular}
\end{table*}

\begin{table*}[ht]
\centering
\caption{Fine-tuned SLM performance compared to the LLM baseline and Phi-3.5 without fine-tuning.}
\label{tab:camphor}
\resizebox{\textwidth}{!}{%
\begin{tabular}{ccccc}
\toprule
\multirow{3}{*}{\textbf{Metric}} & \multicolumn{4}{c}{\textbf{Model}}  \\
\cmidrule(l){2-5}
& \makecell{\textbf{Phi-3.5} \\ (fine-tuning)} & \makecell{\textbf{Gemma-2} \\ (fine-tuning)} & \makecell{\textbf{Claude-3.5} \\ (LLM baseline) } & \makecell{\textbf{Phi-3.5} \\ (no fine-tuning)} \\
\midrule
\textbf{Tool F1} / \% & \cellcolor{gray!20}\textbf{67.84} & 63.72 & 51.31 & 12.94  \\
\textbf{Delexicalized Plan F1} / \% & \cellcolor{gray!20}\textbf{44.85} & 41.57 & 30.07 & 10.39 \\
\textbf{Plan F1} / \% & \cellcolor{gray!20}\textbf{38.77} & 37.43 & 27.96 & 9.72  \\
\bottomrule
\end{tabular}%
}
\end{table*}

\subsection{LLM Baseline Experiments}
Before presenting results for the fine-tuned SLM agents, we first evaluated the performance of state-of-the-art LLMs on the CAMPHOR test set as baselines. We choose Claude-3.5 as the LLM for evaluation to avoid any potential label leakage as the CAMPHOR dataset is generated with GPT-4 in the loop. 

A key difference between instruction-based inference and fine-tuning is that the former relies fully on the prompt instructions which must be clear and often framed with specific structure to guide the pre-trained model. In comparison (as we will show in Section~\ref{exp:slm}), the prompts used in fine-tuning  can be more concise and tailored according to prompt budgets, as the model is tuned to act for certain pattern of inputs. Given the requirement of instruction-based inference, we evaluated a wide range of prompting strategies and aim to pick the best for the comparison with SLM fine-tuning. The prompting strategies include:

% \begin{figure}[t]
%     \centering
%     \includegraphics[width=\columnwidth]{Claude.png} % Use \columnwidth to scale to 1 column
%     \caption{Claude 3.5 Sonnet performance with different prompting strategies.}
%     \label{fig:claude}
% \end{figure}

\begin{itemize}
    \item \textit{Static} employs a consistent prompt template which contains all available function definitions for all CAMPHOR agents. The LLM agent is tasked to generate a sequence of function calls for each CAMPHOR query. The prediction history is also appended to the prompt of each turn. 
    \item \textit{ReAct} is similar to the \textit{Static} baseline, but additionally has the option to perform an explicit reasoning step before generating a function call.
    \item \textit{Reflexion} is similar to ReAct, but additionally incorporates a reflection step to examine the generated function calls and provide feedback. \textit{Reflexion} inherently requires multiple trials, the number of which is set to 3.
    %\item \textit{AUTOACT}: We employ three distinct prompt templates for each agent in our system. In each round, the Plan Agent initially generates a set of thoughts to analyze the observations and subsequently produces a function name (excluding parameters). The Tool Agent then completes the function by supplying the necessary parameters. Once a tentative answer is prepared for submission, the Reflection Agent evaluates it by generating reflective thoughts on its validity, deciding whether the answer is acceptable or if further problem-solving by the agents is required.
    \item \textit{AUTOACT} employs three district prompt templates that respectively handle function call generation, parameter filling and reflection of the results. Similar to other baselines, each CAMPHOR query is parsed as a sequence of function calls with parameter values. The prediction history is also appended to the prompt of each turn. 
    \item \textit{CAMPHOR Agents} employ distinct prompt templates for each CAMPHOR agent. The prompt of each agent contains agent-specific task descriptions, function calls of that agent and in-context examples. The prediction history is also appended to the prompt of each turn.  This setting is closer to the dynamic prompt construction adopted in fine-tuning. Examples for the prompts can be found in the Appendix. 
    %In each round, the High-order reasoning agent first selects a sub-agent and generates corresponding instructions. If the selected sub-agent is either the device information agent or the user perception agent, they directly retrieve information from the environment. For all other sub-agents, which are based on large language models (LLMs), actions are generated first, followed by interaction with the environment.
\end{itemize}

% \begin{figure}[t]
%     \centering
%     \includegraphics[width=\columnwidth]{CAMPHOR.png} % Use \columnwidth to scale to 1 column
%     \caption{Fine-tuned SLM performance Vs. LLM performance.}
%     \label{fig:camphor}
% \end{figure}

%\begin{table*}[ht]
%\centering
%\begin{tabular}{lccccc}
%\toprule
%Prompting Stategy & Baseline & ReAct & Reflexion & AUTOACT & CAMPHOR Agents \\
%\midrule
%\# of Prompting Templates & 1 & 1 & 1 & 3 & 6 \\
%\# of Steps Prompted & 1 & 1 & 1 & 1-2 & 1-2 \\
%Explicit Reasoning & No & Yes & Yes & Yes & Yes \\
%Reflection Module & No & No & Yes & Yes & Yes \\
%\bottomrule
%\end{tabular}
%\caption{Comparison of LLM Prompting Strategies}
%\end{table*}

Evaluation results of various LLM prompting strategies are presented in Table~\ref{tab:claude}. 
%For the Reflexion prompting strategy, we report the performance after three trials. 
The table demonstrates the importance of dynamic, agent-specific prompting strategies in solving CAMPHOR queries. One conclusion is that it is sub-optimal to overload all agent instructions and function definitions into a static prompt template. Given the results, we select \textit{CAMPHOR-agent} as the LLM baseline to compare with fine-tuned SLM agents.
% Add some findings here.

% \begin{figure}[ht]
%     \centering
%     \includegraphics[width=\columnwidth]{GPT-4.png} % Use \columnwidth to scale to 1 column
%     \caption{GPT-4 performance as an agent across various settings.}
%     \label{fig:gpt4}
% \end{figure}

% \begin{figure}[t]
%     \centering
%     \includegraphics[width=\columnwidth]{Claude.png} % Use \columnwidth to scale to 1 column
%     \caption{Claude 3.5 Sonnet performance with different prompting strategies.}
%     \label{fig:claude}
% \end{figure}

%Our results demonstrate that the CAMPHOR framework consistently outperforms all single-agent and multi-agent baselines when applied to GPT-4. For Claude 3.5 Sonnet, CAMPHOR also surpasses the performance of all other baselines, with the exception of Task Completion Call Tool Accuracy.
\begin{table*}[ht]
\centering
\caption{Applying prompt compression significantly reduces the prompt size with marginal changes in accuracy. The prompt size reduction is measured for personal context (PC) and task completion (TC) agents respectively.}
\label{tab:compress}
\begin{tabular}{cccc}
\toprule
\textbf{Metric} & \textbf{No Prompt Compression} & \textbf{Prompt Compression} & \textbf{Relative \(\Delta\)} \\
\midrule
% Response Quality / \% & 82.61 & 82.36 & -0.30\% \\
\textbf{Tool F1} / \% & 66.95 & 65.91 & -1.55\% \\
\textbf{Delexicalized Plan F1} / \% & 44.68 & 44.29 & -0.87\% \\
\textbf{Plan F1} / \% & 39.89 & 38.45 & -3.61\% \\
\midrule
% \textbf{Cost} / \$ per request & 0.0014 & 0.0011 & -21.43\% \\
\textbf{\# of PC Tool Tokens} & 575 & 23 & -96.00\% \\
\textbf{\# of TC Tool Tokens} & 261 & 13 & -95.02\% \\
\bottomrule
\end{tabular}
\end{table*}

\subsection{SLM Fine-tuning Experiments \label{exp:slm}}
Remember that we consider two base SLMs, Phi-3.5 and Gemma-2, for fine tuning. A key question we aim to answer is how to formulate the prompt such that the SLM maintains high accuracy while satisfying the prompt budget of on-device deployments. 

We start with the dynamic prompt formatting function described in Section~\ref{dp}. Each agent prompt contains an agent-specific task description, the prediction history and optionally function descriptions. Compared to the agent-specific prompts in LLM experiments, there are two differences in the SLM fine-tuning. First we only append definitions for the dynamic set of functions for the personal context agent and the task completion agent, since the static functions (and their parameters) can be memorized via fine-tuning. Second, we removed in-context examples for each agent from the prompt, considering the prompt budget and also because the model can be trained to react to input patterns without in-context learning. 

Table~\ref{tab:camphor} shows the results comparing the fine-tuned SLMs with the best LLM prompting strategy. The fine-tuned SLMs, including both Phi-3.5 and Gemma-2, outperform the LLM result in task completion metrics. Meanwhile, the Phi-3.5 model without fine-tuning does poorly in task completion. The results highlight the effectiveness of fine-tuning an SLM for specialized agent tasks, showing it to be more powerful than simply prompting a pre-trained LLM with task instructions. Moreover, the performance of fine-tuning is not compromised by prompt simplification since the model is trained to learn fixed input-output mappings patterns. 
%This may be because the models are explicitly trained to prioritize only the necessary API calls in the gold trajectories, making them more cautious in calling functions, which could lead to a lower PCR recall.

To further optimize the prompt, we remove system instructions from each agent prompt and only reveal the prediction history, based upon which the SLM is fine-tuned to predict the next function call in the trajectory. Surprisingly we found that the prompt simplification leads to only marginal degradation of the task completion, with a plan F1 38.3\% compared to 38.7\% in the original setting. The result demonstrates that fully non-instruction tuning is also a promising direction to further improve on-device efficiency without sacrificing much accuracy.

%To optimize on-device deployment of CAMPHOR, we extract the essential API calls required to generate the final textual response and task completion call for each user request. We then construct the gold trajectories by sequentially incorporating the necessary API calls, ultimately producing the final answer. These trajectories are used to train Small Language Models (SLMs).

% \begin{figure}[t]
%     \centering
%     \includegraphics[width=\columnwidth]{CAMPHOR.png} % Use \columnwidth to scale to 1 column
%     \caption{Fine-tuned SLM performance Vs. LLM performance.}
%     \label{fig:camphor}
% \end{figure}

\subsection{Prompt Compression}
Even though we only append definitions for the dynamic function set in the prompt, they still consume a significant amount of prompt space for large toolboxes. We further experiment with the prompt compression technique described in Section~\ref{pc} where each function definition is represented as a single token in the prompt. 

As shown in Table ~\ref{tab:compress}, applying the prompt compression technique only leads to marginal changes in the task completion F1, from 39.89~\% to 38.45~\%. But it should be noted that the prompt compression technique reduces the number of static prompt tokens (without message history which dynamically grows) further by 96.00\% for the personal context agent and 95.02\% for the task completion agent.

\begin{table*}
\centering
\resizebox{\textwidth}{!}{
\begin{tblr}{
  hlines, vlines,
  colspec = {Q[c,m] Q[c,m] Q[c,m] Q[c,m] Q[c,m]},  
  }
 & \textbf{CAMPHOR LLM}   & \textbf{CAMPHOR SLM}   & \textbf{-instructions}    & \textbf{+compression}  \\
\textbf{Prompt}  & {\{task description\}\\\{agent instructions\}\\\{tool definition\}\\\{message history\}\\\{few-shot examples\}} 
                & {\{agent instructions\}\\\{tool definition\}\\\{message history\}} 
                & {\{tool definition\}\\\{message history\}} 
                & {\{compressed tokens\}\\\{message history\}} \\
{\textbf{Static token}\\\textbf{reduction} \%} 
                & 0\%  & 48.6\%  & 50.3\%  & 98.3\%      
\end{tblr}
}
\caption{A summary of prompt optimisation for CAMPHOR agents. Prompts in SLMs are simplified by virtue of fine-tuning. We further experimented with removing task instructions from the prompt, and compression techniques. Static token reduction rate measures the reduction of static prompt tokens (excluding message history) compared to the prompt used in instruction-based inference.} 
\label{table:prompt} 
\end{table*}

% \begin{table*}
% \centering
% \begin{tblr}{
%   hlines, vlines,
%   }
%  & \textbf{CAMPHOR LLM}   & \textbf{CAMPHOR SLM}   & \quad\quad \textbf{-instructions}    & \quad\quad \textbf{+compression}  \\
% \textbf{Prompt}  & {\{task description\}\\\{agent instructions\}\\\{tool definition\}\\\{few-shot examples\}\\\{message history\}} & {\{agent instructions\}\\\{tool definition\}\\\{message history\}} & {\{tool definition\}\\\{message history\}} & {\{compressed tokens\}\\\{message history\}} \\
% {\textbf{Static token}\\\textbf{reduction} \%} & 0\%  & 48.6\%  & 50.3\%  & 98.3\%                                 
% \end{tblr}
% \caption{A summary of prompt optimisation for CAMPHOR agents. Prompts in SLMs are simplified by virtue of fine-tuning. We further experimented with removing task instructions from the prompt, and compression techniques. Static token reduction rate measures the reduction of static prompt tokens (excluding message history) compared to the prompt used in instruction-based inference.} 
% \label{table:prompt} 
% \end{table*}

Table~\ref{table:prompt} shows a summary of all prompt optimisation steps we have taken for SLM fine-tuning. Comparing to the prompt used for the CAMPHOR LLM baseline, the fine-tuning prompt is made much shorter and concise. We optimised the prompt by removing agent-specific instructions and employing compression technique to represent each function definition as a single token. The total static token reduction rate compared to the baseline is 98.3\%, meaning that the fine-tuned SLM requires much shorter prompts than instruction-based inference in LLM.

\subsection{Comparison with RAG \label{rag}}
\begin{figure}[t]
    \centering
    \includegraphics[width=\columnwidth]{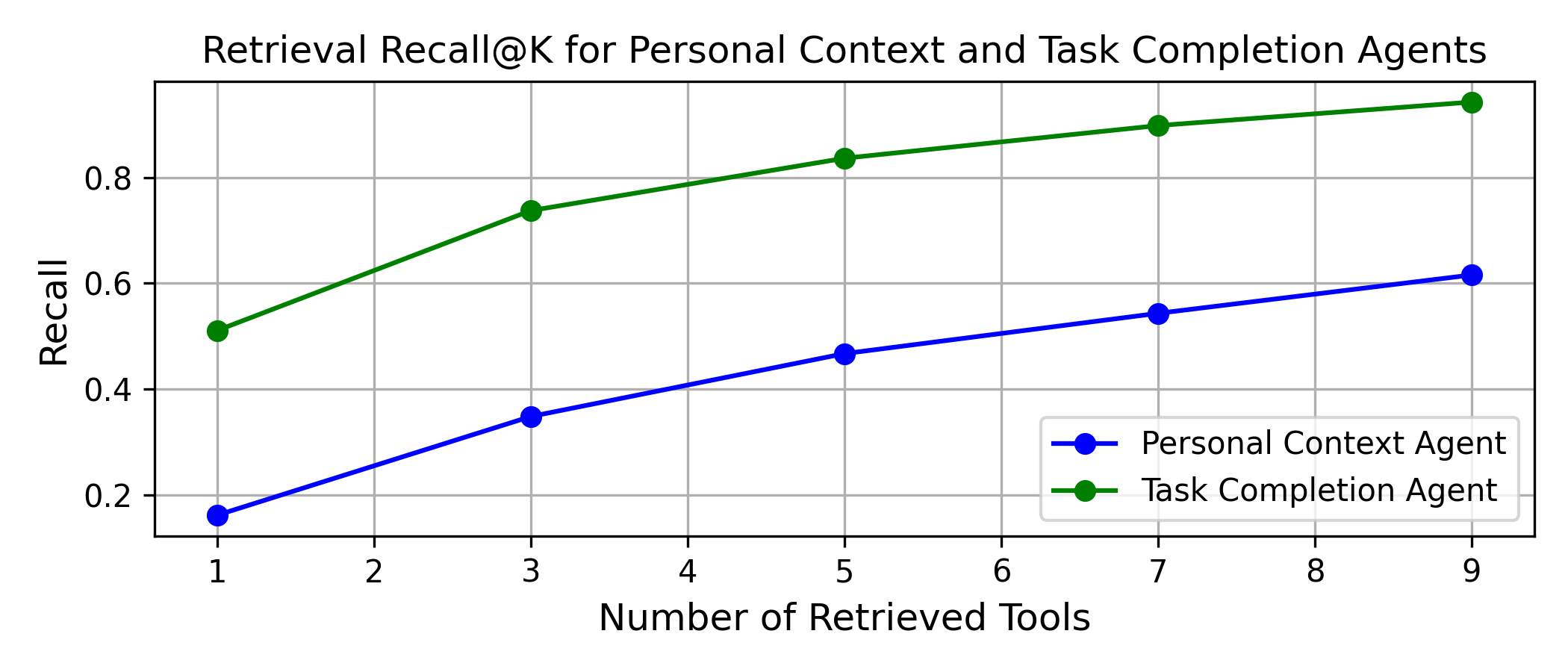} % Use \columnwidth to scale to 1 column
    \caption{Retrieval recall at K computed with an external retrieval model for personal context agent and task completion agent.}
    \label{fig:retrieval}
\end{figure}

One could argue that an alternative approach to generalize to a dynamic toolbox is retrieval-augmented generation (RAG). However, we showcase here that RAG creates a performance bottleneck for the SLM when handling CAMPHOR queries due to sub-optimal retrieval recall. Figure~\ref{fig:retrieval} shows the retrieval recall at K for the CAMPHOR test queries using a Sentence-BERT~\cite{reimers2019sentence} as the retriever. The recall at K=5 is only 0.5 for the personal context agent and 0.8 for the task completion agent. Unsurprisingly the end-to-end plan F1 for a RAG approach with top 5 function definitions in the prompt is only 32.5\% due to error propagation, compared to 38.7\% when all function definitions are present in the prompt. 

\textbf{Why is RAG not working well?} Queries in CAMPHOR are compositional with multiple task completion function calls. The average number of task completion function calls for each query is 3. However, given a tight prompt budget of K=5, it is rather difficult to make sure the retriever is able to fetch all function calls into the K=5 bucket. As a direct consequence, the language model will not see the correct function definitions in the prompt, conditioned on which it is trained to generate the completion. The following shows an example where retrieval failed to find all ground truth:

\begin{verbatim}
Query 
Can you show me the cheapest flight options
to Barcelona next month and add it to my ca-
lendar? Also, let my travel buddy know about
our trip plan.

Ground truth functions 
create_calendar_event,
send_message

Retrieved functions
send_mail, send_message, 
download_appstore_app, play_podcasts, 
create_reminders
\end{verbatim}

While admitting that the performance of the retriever on such queries can be improved with customized fine-tuning and more advanced dense retrieval techniques~\cite{santhanam2022colbertv2}, we contend that the pre-trained language model itself holds greater potential for selecting the appropriate function call in an end-to-end manner, when it has full visibility of the toolbox enabled by prompt compression techniques.

\section{Conclusion}
This work introduces CAMPHOR, a collaborative, SLM-based agent framework designed for personalized query parsing on user devices. CAMPHOR proactively retrieves on-device information and decomposes the understanding tasks into multiple steps of function calls. Our results show that a fine-tuned SLM outperforms instruction-based LLMs in this task. By employing advanced prompt compression techniques, CAMPHOR strikes an optimal balance between accuracy and efficiency, while safeguarding user data directly on the device.

\section*{Limitations}
The personalized user query parsing task studied in this work is restricted to single interactions. While many user queries can indeed be resolved in one interaction, this approach oversimplifies the problem space. In practice, many real-world tasks—especially those requiring user disambiguation or confirmation—still depend on multi-turn interactions between the user and the assistant. In such cases, system policies play a critical role in guiding the conversation and triggering the next agent. Future work should focus on extending CAMPHOR to handle multi-turn conversations, incorporating system policies and user follow-ups.

The simulated device environment in this work also primarily focuses on the "happy path" of personal information retrieval. It does not account for more complex runtime feedback and error-handling logic, such as disambiguation requests for multiple search results, which would need to be communicated back to the user before task continuation. In future, we aim to scale our data simulation approach to handle more complex runtime feedback and in multi-turn conversational settings, as discussed in the first paragraph.

%\section*{Ethics Statement}

%\section*{Acknowledgements}
%This section is intentionally left blank.

\bibliography{custom}

\appendix
% \section{Appendix}
% \label{sec:appendix}

%\section{Data Generation Example}

%Figure ~\ref{fig:example} shows an example data generation process, in which GPT-4o is instructed to annotate the query in a multi-step manner, with each new generation building upon the previous ones.

%\begin{figure*}[ht]
%    \centering
%    \includegraphics[width=\textwidth]{example.png} % Use \columnwidth to scale to 1 column
%    \caption{Example data generation. GPT-4o is instructed to annotate the query in a multi-step manner, with each new generation based on the previous ones.}
%    \label{fig:example}
%\end{figure*}

\section{Fine-tuning details}

Table~\ref{tab:hyperparameters} provides hyper-parameters for fine-tuning the SLMs, including both Phi-3.5 and Gemma-2.

\begin{table*}[ht!]
\centering
\caption{Hyper-parameters for fine-tuning Phi-3.5 and Gemma-2.}
\label{tab:hyperparameters}
\begin{tabular}{lcc}
\toprule
\textbf{Hyper-parameters} & \textbf{Phi-3.5} & \textbf{Gemma-2} \\
\midrule
\textbf{Optimization} & & \\
Batch Size & 128 & 32 \\
Training Steps & 443 & 600 \\
Warmup Ratio & 0.03 & 0.03 \\
LR & $1\times10^{-4}$ & $1\times10^{-4}$ \\
Seed & 42 & 42 \\
\midrule
\textbf{Hardware} & & \\
Data Type & bfloat16 & bfloat16 \\
Accelerator & A100 80G & A100 80G \\
\midrule
\textbf{LoRA Config} & & \\
LoRA Alpha & 16 & 16 \\
LoRA Dropout & 0.05 & 0.05 \\
LoRA Rank & 16 & 16 \\
Target Proj Modules & qkv, o, gate\_up, down & q, k, v, o, gate, up, down \\
\midrule
\textbf{Quantization} & & \\
Quantization Type & nf4 & nf4  \\
Double Quantization & true & true \\
Computation Type & bfloat16 & bfloat16 \\
\bottomrule
\end{tabular}
\end{table*}

\section{Prompts}
The CAMPHOR baseline prompt used to instruct the LLM are shown below. The prompt serves as the starting point to simplify and derive various fine-tuning prompts.

The prompt is constructed from the chat template with a system role and a user role, each of which contains a few variables. We present the templates for the system role and the user role respectively.
\onecolumn

The following are the templates for the system role and user role:
mdframed

\begin{tcolorbox}[colframe=black, colback=gray!10, title=System-role Template, breakable]

\{task description\} (Optional)

\{agent architecture description\} (Optional)

Here are available API calls:

\{tool definition\} (Optional)
\end{tcolorbox}

\begin{tcolorbox}[colframe=black, colback=gray!10, title=User-role Template, breakable]

\{agent specific instruction\} (Optional)

Here is the message history:

\{message history\}

\{few shot examples\} (Optional)
\end{tcolorbox}

The following is the task description in the system role:
\begin{tcolorbox}[colframe=black, colback=gray!10, title=Task Description, breakable]
    You are a helpful digital assistant. An iPhone user has issued a query to you. Your ultimate goal is to provide an accurate and helpful response and complete any related tasks. This may involve utilizing additional context, such as personal contexts and relevant facts, to enhance the user experience.
\end{tcolorbox}

The following are the agent descriptions in the system role:

\begin{tcolorbox}[colframe=black, colback=gray!10, title=Agent Descriptions, breakable]
To successfully complete a complex task, collaboration among the following types of agents is required: \\
1. High Order Reasoning Agent. This agent is used to plan the specific execution process of the task, solving a given task by determining the order in which other expert agents are invoked. Also, this agent will be responsible for overseeing the communication between the expert agents, effectively using their skills to complete sub-tasks. \\
2. Information Agent. This agent is responsible for providing direct information including location information, time information or screen information. 
Location Information: Detailed current location information of user. \\
Time Information: Detailed current time information of user. \\
Screen Information: A detailed textual description of the user's screen content. When calling this agent, please select one type of information to retrieve. \\
3. Perception Agent. This agent is responsible for translating the onscreen context into a high-level understanding of the user's intent. Note that this intent is abstract; if more detailed textual information is needed, the Information Agent would be a better choice. \\
4. Personal Context Retrieval Agent. This agent is responsible for actively seeking relevant personal context that would be helpful in more accurate and personalized response. \\
5. Tool Calling Agent. This agent is responsible for calling useful tools. Tools can include external tools like web searches, Wikipedia, and calculators. \\
6. Answer Agent. This agent is responsible for generating tentative responses and task completion API calls based on the message history. These responses can then be reviewed by Reflection Agent and polished for the final answer. \\
7. Reflection Agent. This agent evaluates the proposed final response and execution history to determine whether the suggested textual response and task completion API calls are appropriate for the given query and provides recommendations for improving the response. \\
8. Response Submit Agent. This agent is similar to Answer Agent, and is also responsible for generating tentative responses and task completion API calls based on the message history. The output from this agent will be directly submitted to the user. \\
These agents will communicate by sending messages and sharing a message history.
\end{tcolorbox}

The following are the tool definitions for each agent in their system role: 

\begin{tcolorbox}[colframe=black, colback=gray!10, title=Tool Definition for Device Information Agent, breakable]
    get\_screen\_information(): Get a detailed textual description of the user's screen content. \\
    get\_location\_information(): Get detailed current location information of user. \\
    get\_time\_information(): Get detailed current time information of user. 
\end{tcolorbox}

\begin{tcolorbox}[colframe=black, colback=gray!10, title=Tool Definition for User Perception Agent, breakable]
    get\_intent(): Get a high-level understanding of the user's intent.
\end{tcolorbox}

\begin{tcolorbox}[colframe=black, colback=gray!10, title=Tool Definition for Personal Context Agent, breakable]
get\_settings\_cellular(): Retrieve user's cellular data usage summary. \\
get\_settings\_notifications(keyword): Retrieve user's notifications containing a specific keyword. \\
get\_health\_records(): Retrieve user's health records. \\
get\_health\_medications(): Retrieve user's medication list. \\
get\_fitness\_summary(): Retrieve user's fitness summary and activity. \\
get\_safari\_history(keyword): Retrieve browsing history of Safari containing a specific keyword. \\
get\_news\_history(keyword): Retrieve browsing history of News containing a specific keyword. \\
get\_podcasts\_history(keyword): Retrieve listening history of Podcasts containing a specific keyword. \\
get\_notes\_content(keyword): Retrieve notes containing a specific keyword. \\
get\_reminders\_content(keyword, time\_range): Retrieve reminders containing a specific keyword or/and within a specific time range. \\
get\_calendar\_event(theme, time\_range): Retrieve calendar events related to a theme or/and within a specified time range. \\
get\_mail\_event(theme, time\_range): Retrieve mail invitation or confirmation for events related to a theme or/and within a specified time range. \\
get\_imessage\_history(keyword): Retrieve chatting history of iMessage containing a specific keyword. \\
get\_music\_playlist(keyword): Retrieve songs in user's music playlist containing a specific keyword. \\
get\_voice\_recording(keyword): Retrieve recordings from the user's voice memos with titles containing a specific keyword. \\
get\_books\_library(): Retrieve user's reading books. \\
get\_contacts\_information(keyword): Retrieve contact information, including person\_id, name, phone\_number, relationship. \\
get\_appstore\_history(): Retrieve the purchase and download history of apps. \\
get\_maps\_places(keyword): Retrieve user's saved places containing a specific keyword. \\
get\_amazon\_information(): Retrieve user's Amazon account information. \\
get\_amazon\_orders(keyword): Retrieve user's Amazon orders containing a specific keyword. \\
get\_instagram\_information(): Retrieve user's Instagram account information. \\
get\_instagram\_post(keyword): Retrieve user's Instagram post containing a specific keyword.
\end{tcolorbox}

\begin{tcolorbox}[colframe=black, colback=gray!10, title=Tool Definition for External Knowledge Agent, breakable]
    search\_safari(query): Perform a search in Safari app using the specified query, which can include searches for information, weather forecasts, available items on Amazon, and other types of information.
\end{tcolorbox}

\begin{tcolorbox}[colframe=black, colback=gray!10, title=Tool Definitions for Task Completion Agent, breakable]
play\_podcasts(title): Play a podcast with the specified title. \\
create\_notes(content): Create a note with the specified content. \\
create\_reminders(time, content): Set a reminder with the specified content at the specified time. \\
create\_calendar\_event(time, event\_title): Create a calendar event with the specified event\_title at the specified time. \\
cancel\_calendar\_event(event\_title): Cancel the calendar event with the specified event\_title. \\
send\_mail(receiver, content): Send an email to the receiver with the specified content. \\
send\_imessage\_message(receiver, content): Send a message to the receiver with the specified content via iMessage. \\
play\_music(title): Play music with the specified title. \\
call\_contacts(person): Call the specified person. \\
download\_appstore\_app(app\_name): Download the specified app. \\
show\_maps\_place(name): Show the location of the specified place in the Maps app. \\
show\_amazon\_item(name): Show the page of the specified item on Amazon. \\
create\_instagram\_post(content): Create a new post with the specified content on Instagram.
\end{tcolorbox}

The following are the agent-specific instructions in the user role:

\begin{tcolorbox}[colframe=black, colback=gray!10, title=Agent Specific Instruction for Device Information Agent, breakable]
Now your task is to generate accurate and helpful API calls to retrieve device information based on the message history.
\end{tcolorbox}

\begin{tcolorbox}[colframe=black, colback=gray!10, title=Agent Specific Instruction for User Perception Agent, breakable]
N/A (Just one tool in the toolbox and there are no parameters; no need to call the model to determine the tool.)
\end{tcolorbox}

\begin{tcolorbox}[colframe=black, colback=gray!10, title=Agent Specific Instruction for Personal Context Agent, breakable]
Now your task is to generate accurate and helpful API calls to retrieve personal context based on the message history.
\end{tcolorbox}

\begin{tcolorbox}[colframe=black, colback=gray!10, title=Agent Specific Instruction for External Knowledge Agent, breakable]
Now your task is to generate accurate and helpful API calls to retrieve relevant facts or public information based on the message history.
\end{tcolorbox}

\begin{tcolorbox}[colframe=black, colback=gray!10, title=Agent Specific Instruction for Task Completion Agent, breakable]
Now your task is to generate accurate and personalized textual response and task completion API calls for user based on the message history.
\end{tcolorbox}

The following is an example for the message history. This history gets updated dynamically after an agent is invoked and the execution result is sent back.

\begin{tcolorbox}[colframe=black, colback=gray!10, title=Example of Message History, breakable]
\texttt{[User]}: Can you show me the cheapest flight options to Barcelona next month and add it to my calendar? Also, let my travel buddy know about our trip plan. \\

\texttt{[High Order Reasoning Agent]: [Device Information Agent]} \\

\texttt{[Device Information Agent]: ['get\_location\_information()']} \\

\texttt{[Execution Result]: [{"latitude": 53.3478, "longitude": -6.2597, "city": "Dublin", "country": "Ireland", "postal\_code": "D01 V902", "formatted\_address": "Ryanair Head Office, Airside Business Park, Swords, Co. Dublin, Ireland"}]} \\

\texttt{[High Order Reasoning Agent]: [Personal Context Agent]} \\

\texttt{[Personal Context Agent]: ["get\_contacts\_information(keyword='travel buddy')"]} \\

\texttt{[Execution Result]: [{"person\_id": "003", "name": "Alice Johnson", "phone\_number": "555-123-4567", "relationship": "Travel Buddy", "is\_self": "False"}]} \\

\texttt{[High Order Reasoning Agent]: [External Knowledge Agent]} \\

\texttt{[External Knowledge Agent]: [search\_safari(query='Cheapest flights from Dublin to Barcelona January 2024')]} \\

\texttt{[Execution Result]: [Cheapest flights from Dublin to Barcelona in January 2024:
- Tuesday, January 7th: €29.99, Departure at 7:00 AM, Arrival at 10:30 AM.
- Thursday, January 16th: €32.50, Departure at 6:45 AM, Arrival at 10:15 AM.
- Friday, January 10th: €31.00, Departure at 8:00 AM, Arrival at 11:30 AM.]} \\

\texttt{[High Order Reasoning Agent]: [Task Completion Agent]} \\

\texttt{[Task Completion Agent]:}\\
Textual Response:\\
Sure! Here are some of the cheapest flight options to Barcelona next month from various sources:\\
From Ryanair:\\
1. Tuesday, January 7th at 7:00 AM, arriving at 10:30 AM - €29.99.\\
2. Thursday, January 16th at 6:45 AM, arriving at 10:15 AM - €32.50.\\
3. Friday, January 10th at 8:00 AM, arriving at 11:30 AM - €31.00.\\
I will add the cheapest flight, which departs on January 7th at 7:00 AM and costs €29.99, to your calendar and notify your travel buddy.\\

Task Completion API Calls:\\
\texttt{["create\_calendar\_event(time='2024-01-07T07:00:00', event\_title='Flight to Barcelona - Departure from Dublin at 7:00 AM')", "send\_imessage\_message(receiver='555-123-4567', content='We have a flight to Barcelona on January 7th at 7:00 AM. Please be ready!')"]}
\end{tcolorbox}

The following is an example of few-shot examples in the user prompt:

\begin{tcolorbox}[colframe=black, colback=gray!10, title=Few Shot Examples, breakable, breakable]
Here is an example output: \\
\texttt{[get\_maps\_places(keyword="Apple"); get\_imessage\_history(keyword="Pear")]} \\

Strictly use only the available API calls and separate each API call by semicolons in a list.
\end{tcolorbox}

The following is an example of the full prompt:
\begin{tcolorbox}[colframe=black, colback=gray!10, title=An example of the Full Prompt, breakable]
You are a helpful digital assistant. An iPhone user has issued a query to you. Your ultimate goal is to provide an accurate and helpful response and complete any related tasks. This may involve utilizing additional context, such as personal contexts and relevant facts, to enhance the user experience. \\

To successfully complete a complex task, collaboration among the following types of agents is required: \\
1. High Order Reasoning Agent. This agent is used to plan the specific execution process of the task, solving a given task by determining the order in which other expert agents are invoked. Also, this agent will be responsible for overseeing the communication between the expert agents, effectively using their skills to complete sub-tasks. \\
2. Information Agent. This agent is responsible for providing direct information including location information, time information or screen information. 
Location Information: Detailed current location information of user. \\
Time Information: Detailed current time information of user. \\
Screen Information: A detailed textual description of the user's screen content. When calling this agent, please select one type of information to retrieve. \\
3. Perception Agent. This agent is responsible for translating the onscreen context into a high-level understanding of the user's intent. Note that this intent is abstract; if more detailed textual information is needed, the Information Agent would be a better choice. \\
4. Personal Context Retrieval Agent. This agent is responsible for actively seeking relevant personal context that would be helpful in more accurate and personalized response. \\
5. Tool Calling Agent. This agent is responsible for calling useful tools. Tools can include external tools like web searches, Wikipedia, and calculators. \\
6. Answer Agent. This agent is responsible for generating tentative responses and task completion API calls based on the message history. These responses can then be reviewed by Reflection Agent and polished for the final answer. \\
7. Reflection Agent. This agent evaluates the proposed final response and execution history to determine whether the suggested textual response and task completion API calls are appropriate for the given query and provides recommendations for improving the response. \\
8. Response Submit Agent. This agent is similar to Answer Agent, and is also responsible for generating tentative responses and task completion API calls based on the message history. The output from this agent will be directly submitted to the user. \\
These agents will communicate by sending messages and sharing a message history. \\

You are the Personal Context Agent that is responsible for actively seeking relevant personal contexts that would be helpful in more accurate and personalized response. The High Order Reasoning Agent has assigned a task to you. Could you please generate a sequence of personal context retrieval API calls to retrieve relevant personal context in various smartphone apps based on the available calls and the message history? \\

Here are available API calls to retrieve relevant personal information for each app: \\
get\_settings\_cellular(): Retrieve user's cellular data usage summary. \\
get\_settings\_notifications(keyword): Retrieve user's notifications containing a specific keyword. \\
get\_health\_records(): Retrieve user's health records. \\
get\_health\_medications(): Retrieve user's medication list. \\
get\_fitness\_summary(): Retrieve user's fitness summary and activity. \\
get\_safari\_history(keyword): Retrieve browsing history of Safari containing a specific keyword. \\
get\_news\_history(keyword): Retrieve browsing history of News containing a specific keyword. \\
get\_podcasts\_history(keyword): Retrieve listening history of Podcasts containing a specific keyword. \\
get\_notes\_content(keyword): Retrieve notes containing a specific keyword. \\
get\_reminders\_content(keyword, time\_range): Retrieve reminders containing a specific keyword or/and within a specific time range. \\
get\_calendar\_event(theme, time\_range): Retrieve calendar events related to a theme or/and within a specified time range. \\
get\_mail\_event(theme, time\_range): Retrieve mail invitation or confirmation for events related to a theme or/and within a specified time range. \\
get\_imessage\_history(keyword): Retrieve chatting history of iMessage containing a specific keyword. \\
get\_music\_playlist(keyword): Retrieve songs in user's music playlist containing a specific keyword. \\
get\_voice\_recording(keyword): Retrieve recordings from the user's voice memos with titles containing a specific keyword. \\
get\_books\_library(): Retrieve user's reading books. \\
get\_contacts\_information(keyword): Retrieve contact information, including person\_id, name, phone\_number, relationship. \\
get\_appstore\_history(): Retrieve the purchase and download history of apps. \\
get\_maps\_places(keyword): Retrieve user's saved places containing a specific keyword. \\
get\_amazon\_information(): Retrieve user's Amazon account information. \\
get\_amazon\_orders(keyword): Retrieve user's Amazon orders containing a specific keyword. \\
get\_instagram\_information(): Retrieve user's Instagram account information. \\
get\_instagram\_post(keyword): Retrieve user's Instagram post containing a specific keyword. \\

Now your task is to generate accurate and helpful API calls to retrieve personal context based on the message history. \\

Here is the message history: \\

\texttt{[User]}: Can you show me the cheapest flight options to Barcelona next month and add it to my calendar? Also, let my travel buddy know about our trip plan. \\

\texttt{[High Order Reasoning Agent]: [Device Information Agent]} \\

\texttt{[Device Information Agent]: ['get\_location\_information()']} \\

\texttt{[Execution Result]: [{"latitude": 53.3478, "longitude": -6.2597, "city": "Dublin", "country": "Ireland", "postal\_code": "D01 V902", "formatted\_address": "Ryanair Head Office, Airside Business Park, Swords, Co. Dublin, Ireland"}]} \\

\texttt{[High Order Reasoning Agent]: [Personal Context Agent]} \\

Here is an example output: \\
\texttt{[get\_maps\_places(keyword="Apple"); get\_imessage\_history(keyword="Pear")]} \\

Strictly use only the available API calls and separate each API call by semicolons in a list.
\end{tcolorbox}

\twocolumn

\twocolumn
\end{document}